\pdfoutput=1

\documentclass[11pt]{article}
\usepackage[dvipsnames]{xcolor}
\usepackage{float}

\usepackage[]{acl2023}

\usepackage[utf8]{inputenc}
\usepackage{pgfplots}
\usepackage{inconsolata}
\usepackage{booktabs}
\usepackage{multirow}
\usepackage{amsmath}
\usepackage{amsfonts}
\usepackage{amsthm}
\usepackage{pifont}
\usepackage{xspace}
\usepackage{graphicx}
\usepackage{bm}
\usepackage{xcolor, colortbl}
\usepackage[most]{tcolorbox}
\usepackage{csquotes}
\usepackage{anyfontsize}
\usepackage{comment}
\usepackage{float}
\usepackage{cuted}
\usepackage{tikz}
\usepackage{listings,multicol}
\usepackage{filecontents}
\usepackage{dsfont}
\DeclareUnicodeCharacter{2212}{−}
\usepgfplotslibrary{groupplots,dateplot}
\usetikzlibrary{patterns,shapes.arrows}
\pgfplotsset{compat=newest}

\usepackage{tikzscale}
\usepackage{amsmath}
\newcommand{\probP}{\text{I\kern-0.15em P}}

\newcommand{\colorBoxColor}{SkyBlue}

\newcommand{\contrast}{\mathcal{D}_{contr}}

\newcommand{\choices}{\texttt{Choices:$\backslash$n(A)} $c_a$ \texttt{$\backslash$n(B)} $c_b$ \texttt{$\backslash$n(C)} $c_c$ \texttt{$\backslash$n(D)} $c_d$ \\}

\usepackage{multirow, colortbl}

\usepackage{tabularx,booktabs}
\usepackage{makecell}

\usepackage[normalem]{ulem}
\useunder{\uline}{\ul}{}

\usepackage{graphicx}

\definecolor{ablation6}{HTML}{fcefed}
\definecolor{ablation_tie}{HTML}{fce3e1}

\definecolor{ablation5}{HTML}{fcd8d4}
\definecolor{ablation4}{HTML}{FBC3BC}
\definecolor{ablation3}{HTML}{F7A399}
\definecolor{ablation2}{HTML}{F38375}
\definecolor{ablation1}{HTML}{EF6351}
\definecolor{UMDred}{HTML}{ed1c24}
\definecolor{UMDyellow}{HTML}{ffc20e}
\definecolor{CustomGreen}{HTML}{1FC801}

\definecolor{bggray}{rgb}{0.95, 0.95, 0.95}
\usepackage[%
    framemethod=tikz,
    skipbelow=\topskip,
    skipabove=\topskip
]{mdframed}
\mdfsetup{%
    leftmargin=0pt,
    rightmargin=0pt,
    backgroundcolor=bggray,
    middlelinecolor=black,
    roundcorner=3
}

\newtcolorbox[list inside=prompt,auto counter,number within=section]{prompt}[1][]{
    colbacktitle=black!60,
    fonttitle=\small,
    coltitle=white,
    fontupper=\footnotesize,
    boxsep=4pt,
    left=0pt,
    right=0pt,
    top=0pt,
    bottom=0pt,
    boxrule=1pt,
    #1,
}

\usepackage[]{algpseudocode}
\usepackage[]{algorithm}
\usepackage{float}
\algtext*{EndFor}%
\algtext*{EndProcedure}%

\usepackage{booktabs}
\usepackage[normalem]{ulem}
\useunder{\uline}{\ul}{}

\usepackage{times}
\usepackage{latexsym}
\usepackage{adjustbox}

\usepackage[T1]{fontenc}
\usepackage{amsmath}
\usepackage{amssymb}
\usepackage{booktabs}
\usepackage{multirow}
\usepackage{scalerel,xparse}
\usepackage[utf8]{inputenc}

\usepackage{cleveref}
\usepackage{microtype}
\usepackage{enumitem}
\crefformat{section}{\S#2#1#3}
\crefformat{subsection}{\S#2#1#3}
\crefformat{subsubsection}{\S#2#1#3}

\title{Is Your Large Language Model Knowledgeable or a Choices-Only Cheater?}

\author{Nishant Balepur \\
  University of Maryland\\
  \texttt{nbalepur@umd.edu} \\\And
  Rachel Rudinger \\
  University of Maryland \\
  \texttt{rudinger@umd.edu} \\}

\begin{document}
\maketitle

\begin{abstract} {
Recent work shows that large language models (LLMs) can answer multiple-choice questions using only the choices, but does this mean that MCQA leaderboard rankings of LLMs are largely influenced by abilities in choices-only settings? To answer this, we use a contrast set that probes if LLMs over-rely on choices-only shortcuts in MCQA. While previous works build contrast sets via expensive human annotations or model-generated data which can be biased, we employ graph mining to extract contrast sets from existing MCQA datasets. 
We use our method on UnifiedQA, a group of six commonsense reasoning datasets with high choices-only accuracy, to build an 820-question contrast set. After validating our contrast set, we test 12 LLMs, finding that these models do not exhibit reliance on choice-only shortcuts when given both the question and choices. Thus, despite the susceptibility~of MCQA to high choices-only accuracy, we argue that LLMs are not obtaining high ranks on MCQA leaderboards just due to their ability to exploit choices-only shortcuts.\footnote{Our code is available at \url{https://github.com/nbalepur/mcqa-artifacts}}

}
\end{abstract}

\section{Introduction}

Multiple-choice question answering (MCQA) is a popular task to test the knowledge of large language models (LLMs) \cite{robinson2023leveraging}. However, recent work shows that LLMs surpass majority class baselines in choices-only settings---when no question and just the choices are given in a prompt \cite{balepur2024artifacts}. This raises the question: \textit{Do models obtain high ranks in MCQA leaderboards due to their pretraining knowledge or their ability to exploit choices-only shortcuts?} Resolving this query is key to ensure that MCQA leaderboards reliably rank the knowledge of LLMs.

\begin{figure}
    \centering
    \includegraphics[width=0.98\linewidth]{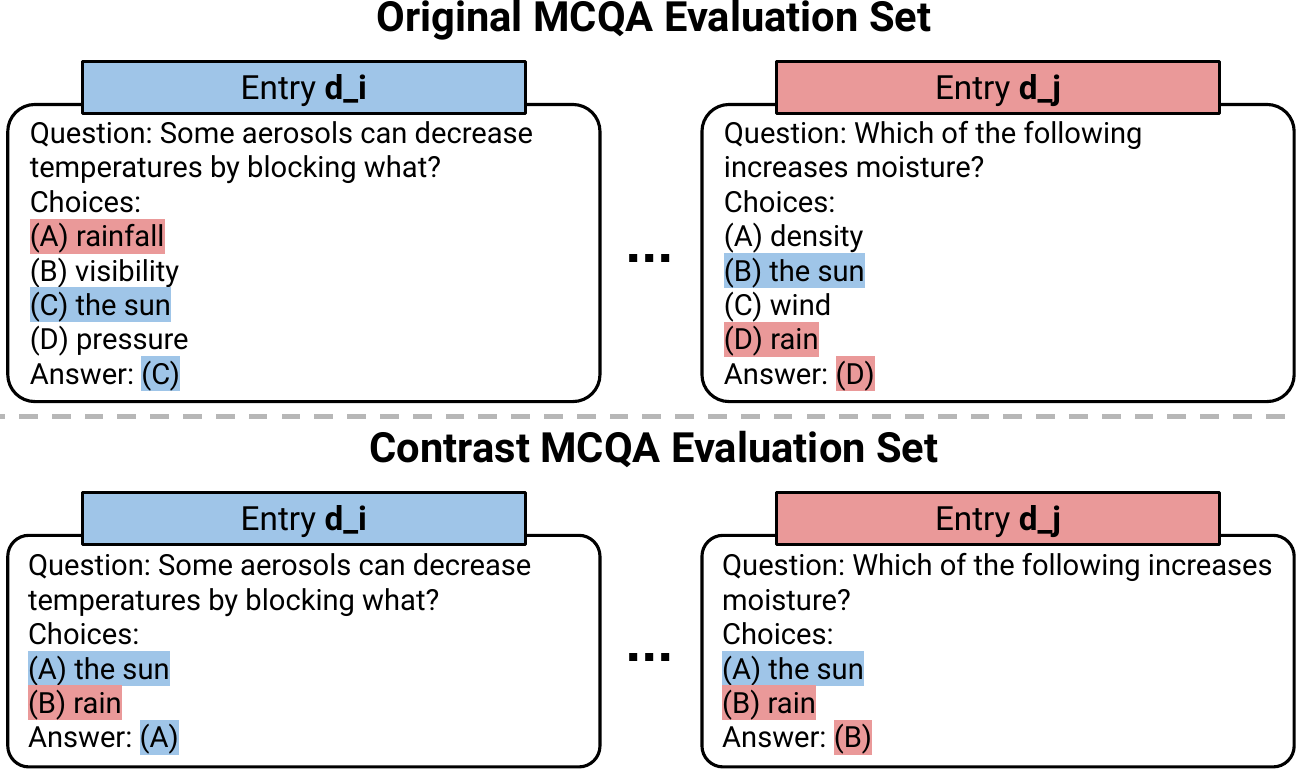}
    \caption{Example of a contrast MCQA evaluation set.}
    \label{fig:intro}
\end{figure}

To answer this question, we use a variation of \emph{contrast sets}---small datasets that test if models ``pay attention'' to perturbed attributes that should alter the model's decision \cite{levesque2012winograd, gardner-etal-2020-evaluating}.
For our purposes, we need a contrast set containing pairs of MC entries with identical answer choices, but varied questions that lead to distinct answers. 
For example, in Figure~\ref{fig:intro} (bottom), the MC entries $d_i$ and $d_j$ have the same choices of ``the sun'' and ``the rain'', but $d_i$ has a question that answered by ``the sun'', and similarly for $d_j$. This design ensures that LLMs relying only on shortcuts or patterns in the choices, while ignoring questions, can perform no better than random chance. 
Thus, if a model ranks highly on an MCQA dataset but largely drops in rank on a contrast set based on this dataset, it would reveal that this model obtains a high rank on the original dataset primarily by employing choices-only shortcuts. 

Contrast sets are usually built through manual annotation efforts \cite{gardner-etal-2020-evaluating, srikanth-rudinger-2022-partial}, as model-generated data can be biased. However, writing MC questions with high-quality distractors is difficult even for experts \cite{gierl2017developing}. 
Further, rewritten questions can exhibit distributional differences from the original questions, altering the difficulty of the MCQA task.

To address this issue, we cast the creation of contrast sets for MCQA datasets to a graph mining task.~We treat each MC entry $d_i$ in the dataset as a vertex~in an undirected graph, and draw edges between entries $d_i$ and $d_j$ if the gold answer from $d_i$ is semantically equivalent to a distractor in $d_j$, and vice versa. For instance, in Figure~\ref{fig:intro} (top), the gold answer of ``rain'' in $d_i$ is semantically similar to the distractor of ``rainfall'' in $d_j$ and vice versa, so we draw an edge between $d_i$ and $d_j$. Thus, an edge ($d_i$, $d_j$) means that the gold answers in $d_i$ and $d_j$ can form a set of choices, with questions in $d_i$ and $d_j$ leading to distinct answers in said choices, mirroring the desired format of our contrast set. We find the maximum matching of this graph to obtain the largest contrast set of distinct MCQA questions derived from the initial dataset. This method overcomes the burden of writing contrast sets, while only minimally using models for semantic equivalence, reducing the risk of model-generated biases. 

We use our approach to build an 820-question contrast set from six commonsense MCQA datasets from the UnifiedQA collection \cite{khashabi-etal-2020-unifiedqa}. We first ask three annotators to assess our contrast set, finding that it has questions with plausible distractors (\cref{subsection:qual}).
This finding suggests that we have built a high-quality MCQA contrast set.

After verifying the quality of our contrast set, we test 12 LLMs \cite{touvron2023llama, penedo2023refinedweb, jiang2023mistral, young2024yi, team2024gemma} on the UnifiedQA evaluation set and its mined contrast set (\cref{subsection:benchmark}). Our LLMs surpass random guessing using just the choices on the original evaluation set, aligning with prior work. \cite{balepur2024artifacts}. However, when prompted with both the question \emph{and} choices, LLM accuracy rankings between the initial evaluation set and contrast set are highly consistent, with Kendall's $\tau$ near $0.9$. 

Since no LLM rank drops markedly, we claim that our tested LLMs are not ranking highly on MCQA leaderboards solely due to their ability to exploit choices-only shortcuts.  
Thus, despite the susceptibility of MCQA to high choices-only accuracy, the task may still reliably rank LLM knowledge.
As a result, we recommend that future works continue to explore the behavior of LLMs in choices-only settings to help explain how LLMs can adeptly perform MCQA without the question. 

\section{Automatic Contrast Set Creation} \label{section:algo}

We assume we are given an MCQA dataset $\mathcal{D}$ with data entries $d_i = (q_i, \mathcal{C}_i, a_i)$, where $q_i$ is a question, $\mathcal{C}_i$ is a list of choices, and $a_i \in \mathcal{C}_i$ is the gold answer. Our goal is to build a contrast set $\contrast$ from $\mathcal{D}$ to probe if LLMs rely on choice-only shortcuts in MCQA. Typically, humans manually create contrast sets \cite{srikanth-rudinger-2022-partial, gardner-etal-2020-evaluating}, as model-generated data can be biased \cite{yu2024large}. However, writing MCQA problems is challenging even for experts \cite{offerijns2020better, gierl2017developing}. Thus, we seek to automatically mine a contrast set $\contrast$ from the original dataset $\mathcal{D}$ without model-generated data. 

To automatic build contrast sets, we need MCQA entry pairs in the style of Figure~\ref{fig:intro} (bottom)---pairs with the same choices $\mathcal{C}' = \{a_i, a_j\}$, but questions $q_i$ and $q_j$ leading to distinct answers $a_i$ and $a_j$ in $\mathcal{C}'$, respectively. We define this format as an \textbf{entry pair} $p_{ij} = \langle (q_i, \{a_i, a_j\}, a_i), (q_j, \{a_i, a_j\}, a_j) \rangle$. Thus, creating the largest possible $\contrast$ with distinct questions is equivalent to finding the maximum set of unique entry pairs $p_{ij}$ in $\mathcal{D}$. In the next sections, we outline our graph-based approach to mine entry pairs from the original dataset $\mathcal{D}$ to form $\contrast$.

\subsection{Graph Representation} \label{subsection:graph_repr}

While a simple strategy to find an entry pair $p_{ij}$ is to sample two entries $(q_i, \mathcal{C}_i, a_i), (q_j, \mathcal{C}_j, a_j) \in \mathcal{D}$ and let $\mathcal{C}' = \{a_i, a_j\}$, this may result in low-quality questions, as there is no constraint that $a_x$ and $a_y$ form a plausible set of choices (\cref{subsection:qual}). For instance, if $a_i$ is a ratio and $a_j$ is an integer, choices $\{a_i, a_j\}$ are implausible and result in a low-quality question. To address this, we intuit that the original dataset $\mathcal{D}$ reveals if two answers $a_i$ and $a_j$ are plausible distractors for each other. For answers $a_i \in \mathcal{C}_x$ and $a_j \in \mathcal{C}_y$, if $a_i$ is semantically equivalent to a distractor $c \in \mathcal{C}_j \setminus \{a_j\}$ and likewise for $a_j$ and $\mathcal{C}_i$, the set of choices $\mathcal{C}' = \{a_i, a_j\}$ will be plausible. 

To execute this idea, we represent the dataset $\mathcal{D}$ as an undirected graph $\mathcal{G}$. Each entry $d_i \in \mathcal{D}$ is a vertex for $\mathcal{G}$. We draw an edge between entries $d_i$ and $d_j$ if the gold answer $a_i$ is semantically equivalent to a distractor $c \in \mathcal{C}_j \setminus \{ a_j \}$ and vice versa, meaning that the choices $a_i$ and $a_j$ can form a plausible set of choices based on $\mathcal{D}$. We create edges with semantic equivalence over exact match to consider choices with minor differences, like ``rain'' and ``rainfall'' in Figure~\ref{fig:intro}, increasing the candidate size of our contrast set. We compute semantic similarity via NLI-based embeddings \cite{conneau-etal-2017-supervised} and set a strict cosine similarity threshold of 0.85 to determine semantic equivalence. 


\subsection{Mining Entry Pairs} \label{subsection:graph_mining}

We now mine entry pairs from the graph $\mathcal{G}$ to build a contrast set $\contrast$. For any edge $(d_i, d_j)$ in $\mathcal{G}$, we know that $a_i$ and $a_j$ form a set of plausible choices. Thus, when $\mathcal{C}' = \{a_i, a_j\}$, entries $(q_i, \mathcal{C}', a_i)$ and $(q_j, \mathcal{C}', a_j)$ form an entry pair $p_{ij}$. Using this idea, we build $\contrast$ by finding the maximum matching \cite{boppana1992approximating} of $\mathcal{G}$, which gives the largest set of edges in $\mathcal{G}$ where no two edges are adjacent. Each edge in the maximum matching form an entry pair for the contrast set and since no edges are adjacent, each entry pair contains two unique questions. This results in the largest possible contrast set without duplicate questions.

\section{Experimental Setup}
\begin{figure}
    \centering
    \includegraphics[width=\linewidth]{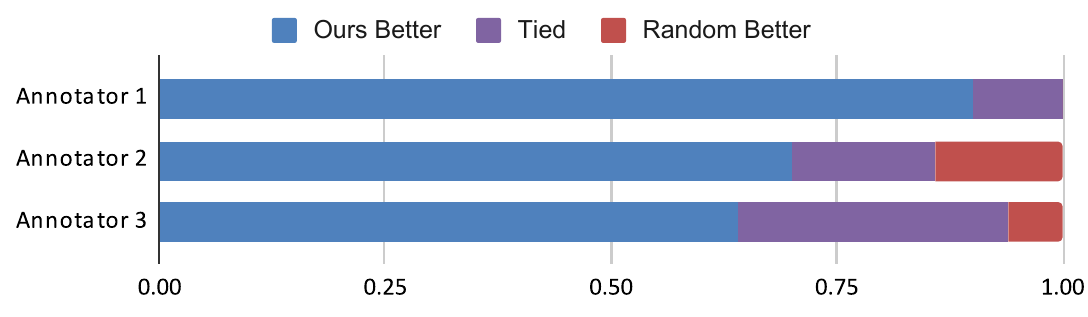}
    \caption{Distractor plausibility ratings across methods.}
    \label{fig:qual}
\end{figure}
\subsection{A Contrast Set for UnifiedQA} \label{subsection:application}

The purpose of our contrast set $\contrast$ is to study whether high choices-only accuracy influences the ranking of LLMs on MCQA leaderboards. Thus, $\contrast$ must be based on a dataset with high accuracy in choices-only settings. The two datasets from \citet{balepur2024artifacts} with the highest choices-only accuracy are commonsense datasets \cite{clark2018think, Zellers2019HellaSwagCA}. Thus, we derive $\contrast$ from an MCQA split of UnifiedQA \cite{khashabi-etal-2020-unifiedqa}, which has 7611 questions from six commonsense datasets: ARC \cite{clark2018think}, OpenBookQA \cite{mihaylov-etal-2018-suit}, CommonsenseQA \cite{talmor-etal-2019-commonsenseqa}, QASC \cite{khot2020qasc}, PIQA \cite{bisk2020piqa}, and SIQA \cite{sap-etal-2019-social}. Using our graph mining algorithm (\cref{section:algo}), we build an 820-question contrast set. This size aligns with contrast set sizes in prior works, ranging from 600 to 1000 \cite{srikanth-rudinger-2022-partial, gardner-etal-2020-evaluating}.

\subsection{Prompt Design}

LLMs are only known to best random guessing with just choices in few-shot prompts. Thus, we follow the few-shot format of \citet{balepur2024artifacts} and use a full prompt (\ref{prompt:full_prompt}) to assess LLMs when they can see both the questions and choices, and a choices-only (\ref{prompt:choices_only_prompt}) prompt for just the choices:

\begin{prompt}[title={Prompt \thetcbcounter: Full Prompt}, label=prompt:full_prompt]
\texttt{Question:} $q$ \\
\choices
\texttt{Answer:} \colorbox{\colorBoxColor}{$a$}
\end{prompt}
\begin{prompt}[title={Prompt \thetcbcounter: Choices-Only Prompt}, label=prompt:choices_only_prompt]
\choices
\texttt{Answer:} \colorbox{\colorBoxColor}{$a$}
\end{prompt}

In the boxes above, the non-highlighted text represents the model input, while the \colorbox{\colorBoxColor}{highlighted text} represents the model generation. In the few-shot prompts, exemplars follow the same format shown in the prompt box with the highlighted text replaced by the ground truth (Example in Appendix~\ref{appendix:prompt_box_example}).

\section{Results}


\subsection{Qualitative Analysis} \label{subsection:qual}

To assess the quality of the contrast set produced by our graph mining algorithm, we ask three Ph.D. students in computer science to compare 50 of our questions versus a baseline that randomly picks entry pairs (details in Appendix~\ref{appendix:qual_details}). These methods only differ by distractors, so following \citet{gierl2017developing}, we ask annotators to compare the \textbf{plausibility} of the two distractors as a proxy for question quality. All three annotators find that our method has significantly more plausible distractors than the baseline (Figure~\ref{fig:qual}), suggesting that our extracted contrast set from UnifiedQA is high-quality.

\begin{figure*}
    \centering
    \includegraphics[width=\linewidth]{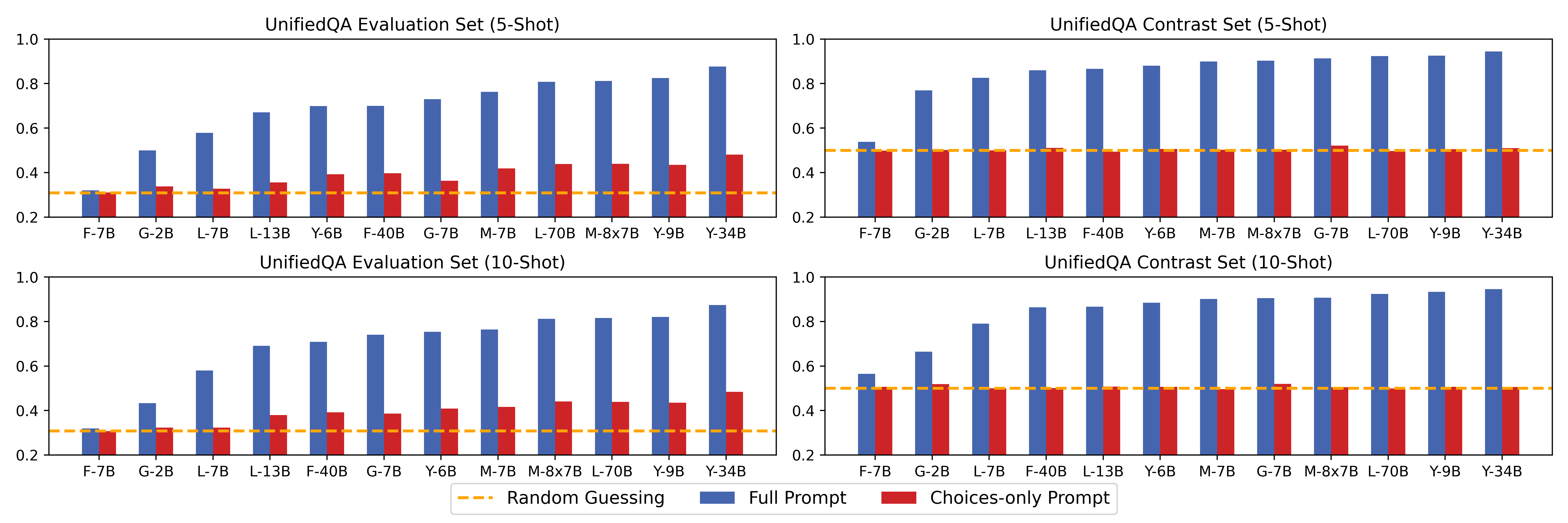}
    \caption{Accuracy of twelve LLMs on the UnifiedQA evaluation set (left) versus its contrast set (right), sorted by full prompt accuracy. We show 5-shot (top) and 10-shot (bottom) prompts, with 3-shot prompts in Appendix~\ref{appendix_3_shot}.}
    \label{fig:quant}
\end{figure*}




\subsection{Are LLMs Knowledgeable or Choices-only Cheaters?} \label{subsection:benchmark}

Following our quality checks, we use our contrast set to study if high choices-only accuracy influences the ranking of LLMs when questions \textit{and} choices are given. We assess 6 LLM families on the UnifiedQA evaluation set and our contrast set: LLaMA-2 \cite{touvron2023llama}, Falcon \cite{penedo2023refinedweb}, Mistral \cite{jiang2023mistral}, Mixtral \cite{jiang2024mixtral}, Gemma \cite{team2024gemma}, and Yi \cite{young2024yi}. We use 5-shot and 10-shot full and choices-only prompts (Prompts~\ref{prompt:full_prompt}, \ref{prompt:choices_only_prompt}). Appendix~\ref{appendix:prompt} has more prompting details.

On the UnifiedQA evaluation set, our LLMs often surpass random guessing with choices-only prompts (Figure~\ref{fig:quant}, left), aligning with prior work \cite{balepur2024artifacts}. Further, LLMs with higher ranks on the UnifiedQA evaluation set using the full prompt tend to have higher accuracy when using the choices-only prompt, suggesting a correlation between an LLM's MCQA leaderboard rank and its ability to exploit choices-only shortcuts. Simply subtracting these values cannot quantify how an LLM performs in MCQA without choices-only shortcuts, since if an LLM can answer a MC question without the question, it does not imply the model is ignoring the question when it has access to the question \cite{srikanth-rudinger-2022-partial}.

Thus, to better quantify if LLMs are obtaining high ranks on UnifiedQA due to their ability to exploit choices-only shortcuts, we compare model ranks on the original UnifiedQA evaluation set to its contrast set. We note that if a certain LLM relied on choice-only shortcuts substantially more than other models, its contrast set accuracy ranking would largely drop compared to its evaluation set accuracy ranking, as it would be penalized for ignoring the question. However, in the UnifiedQA evaluation set and its contrast set, model rankings of full prompt accuracy are consistent; the 5-shot and 10-shot rankings have Kendall's $\tau$ of 0.88 and 0.91, indicating high consistency. Thus, we claim that the MCQA rankings of our LLMs on UnifiedQA do not primarily stem from their ability to perform well in choices-only settings, and none of our models are considered ``choices-only cheaters.''

We find that if an LLM succeeds with choices-only prompts, it does not imply that this model's performance in MCQA solely stems from its choices-only abilities.
As a result, we believe that despite high choices-only accuracy, MCQA may still be a reliable task to rank the knowledge of LLMs.
Further, our results stress the need for more work in explaining how high choices-only accuracy occurs. 
We believe such efforts are crucial to better interpret LLM knowledge and decision-making.

\section{Related Work}

\noindent \textbf{Contrast Sets:} Contrast sets \cite{gardner-etal-2020-evaluating} or counterfactual augmentations \cite{Kaushik2020Learning, srikanth2024often}, are datasets that probe if models ``pay attention'' to desired attributes \cite{elazar2023measuring}. This technique has been applied to many tasks, including natural language inference \cite{glockner-etal-2018-breaking, ribeiro-etal-2020-beyond}, story generation \cite{qin-etal-2019-counterfactual}, and ethical judgements \cite{hendrycks2021aligning}. While these datasets are often created manually, many works use generation models \cite{wu-etal-2021-polyjuice, fryer-etal-2022-flexible} to create contrast sets. Instead, we are the first to employ graph mining to build contrast sets, limiting the potential for model-generated biases.

\noindent \textbf{MCQA Evaluation:} MCQA is a popular testbed not only for benchmarking LLMs \cite{open-llm-leaderboard, lee2023holistic}, but also for interpreting LLM decision-making. Previous works use MCQA to study prompt sensitivity \cite{pezeshkpour2023large, zheng2024large}, logical robustness \cite{balepur2023s}, and recently, the ability to perform MCQA without using the question \cite{balepur2024artifacts}. We give more insights into this last phenomenon by probing if LLMs ignore the question even when it is given in the prompt.

\section{Conclusion}

We find that while LLMs can perform well in MCQA without access to the question, it does not mean that model rankings on MCQA leaderboards are largely influenced by this ability. 
This result supports the claim that MCQA can rank the knowledge and ability of LLMs to reason over both questions and choices. 
Further, we are aligned with recent work that suggests that high choices-only accuracy does not necessarily imply that models are incapable of true reasoning or comprehension, so we hope future works continue to explore what strategies LLMs may employ to perform well in choices-only settings. 
Our application of graph mining to MCQA sheds light on one way to do this---the automatic construction of contrast sets---and we hope similar methods can be applied to other tasks to enhance LLM interpretability.

\section{Limitations}

One limitation lies in the application of our graph mining algorithm solely to the UnifiedQA dataset collection. We choose UnifiedQA for its tendency to elicit high accuracy with choices-only prompts, as commonsense reasoning MCQA datasets have shown to be susceptible to this phenomenon. Since our results show that LLMs rankings are highly consistent on this dataset prone to high choices-only accuracy, we believe these findings will hold for other MCQA datasets like MMLU \cite{Hendrycks2020MeasuringMM} with lower choices-only accuracy. However, we invite future research to apply our graph mining algorithm to other datasets, including non-MCQA datasets, to build contrast sets that can further probe LLM decision-making.

Further, we acknowledge that our contrast set contains MCQA questions limited to two choices, diverging from the original evaluation set's range of two to eight choices. While having less options does make it more likely for a model to guess the right answer, our qualitative analysis shows that the concepts tested in our contrast set are not markedly different in plausibility (\cref{subsection:qual}), and thus are not too easy. Further, while our contrast set is easier in theory, it still preserves LLM rankings, even on the subset used to derive the contrast set (Appendix~\ref{eval_subset}), ultimately supporting the idea that MCQA can reliably rank LLMs capabilities.

\section{Ethical Considerations}

When models heavily rely on patterns or biases present in datasets, we may overestimate model abilities and face generalizability issues during deployment. In this work, we probe the extent to which LLMs over-rely on patterns in MCQA choices when provided both the question and choices in the prompt, ultimately finding that this effect is small. However, we believe it is still critical for LLM practitioners to be aware that LLMs can outperform random guessing when using just the choices as input, as this could have downstream effects. Thus, we encourage future research efforts in designing special datasets that can help interpret specific abilities within LLM decision-making. 

Further, we note that when any model is used in a data creation pipeline, there is the possibility of models propagating their own biases. We specifically address this issue by designing a graph mining algorithm that leverages minimal model intervention, only in the form of computing semantic similarity, which greatly lowers this risk compared to synthetic data generators like LLMs. We hope future works can adopt data creation pipelines with minimal model use similar to ours to avoid the risk of generating model-specific biases or artifacts. 

\section{Acknowledgements}

We would like to thank members of the CLIP lab at the University of Maryland and external collaborators for their feedback and discussions of this work, including Neha Srikanth, Shi Feng, and Jordan Boyd-Graber.
We also thank Yu Hou and Dayeon Ki for their help with annotations. 
This material is based upon work supported by the
National Science Foundation Graduate Research Fellowship Program under Grant No. DGE 2236417. Any opinions, findings, and conclusions or recommendations expressed in this material
are those of the author(s) and do not necessarily reflect the views of the National Science Foundation.

\bibliography{custom}
\bibliographystyle{acl_natbib}

\appendix \label{sec:appendix}

\clearpage

\section{Experimental Setup}

\subsection{Dataset Details} \label{appendix:dataset_details}

The UnifiedQA evaluation set has questions from the evaluation sets of the following six datasets:

\begin{itemize}
    \item \textbf{ARC:} 1172 four-choice questions drawn from grade-school science questions.
    \item \textbf{OpenBookQA:} 500 four-choices questions modeled after open-book exams.
    \item \textbf{QASC:} 926 eight-choice questions about grade school science with a focus on sentence composition.
    \item \textbf{CommonsenseQA:} 1221 four-choice questions meant to test commonsense knowledge from ConceptNet.
    \item \textbf{Physical IQa:} 1838 two-choice questions about physical commonsense reasoning.
    \item \textbf{Social IQA:} 1954 three-choice questions involving reasoning about everyday social interactions.
\end{itemize}

After running our algorithm, our contrast set contains 377 questions from CommonsenseQA, 285 questions from QASC, 79 questions from ARC, 53 questions from Social IQa, 22 questions from OpenBookQA, and 4 questions from Physical IQa, all of which have two choices.

\subsection{Prompt Box Example} \label{appendix:prompt_box_example}

The following subsection is adapted directly from the Appendix of \citet{balepur2024artifacts} to highlight the utility of their prompt boxes. 

Below, we provide a detailed example to illustrate the application of our prompt boxes. Suppose we have the full prompt (Prompt~\ref{prompt:full_prompt}):

\begin{prompt}[title={Prompt 2.1: Full Prompt}, label=prompt:full_prompt_appendix]
\texttt{Question:} $q$ \\
\texttt{Choices:} $\mathcal{C}$ \\
\texttt{Answer:} \colorbox{\colorBoxColor}{$a$}
\end{prompt}

In the above prompt, the LLM uses the question $q$ and choices $\mathcal{C}$ as input and is asked to generate the letter of the answer $a$. Suppose we have 5 few-shot examples, with questions $q_1$ ,..., $q_5$, list of choices $\mathcal{C}_1$, ..., $\mathcal{C}_5$, and ground truth answers $a_1$, ..., $a_5$. The expanded few-shot prompt for the prompt box is written below:

\begin{prompt}[title={Prompt 2.1: Full Prompt Expanded}, label=prompt:full_prompt_appendix_expanded]
\texttt{Question:} $q_1$ \\
\texttt{Choices:} $\mathcal{C}_1$ \\
\texttt{Answer:} $a_1$ \\

\texttt{Question:} $q_2$ \\
\texttt{Choices:} $\mathcal{C}_2$ \\
\texttt{Answer:} $a_2$ \\

\texttt{Question:} $q_3$ \\
\texttt{Choices:} $\mathcal{C}_3$ \\
\texttt{Answer:} $a_3$ \\

\texttt{Question:} $q_4$ \\
\texttt{Choices:} $\mathcal{C}_4$ \\
\texttt{Answer:} $a_4$ \\

\texttt{Question:} $q_5$ \\
\texttt{Choices:} $\mathcal{C}_5$ \\
\texttt{Answer:} $a_5$ \\

\texttt{Question:} $q$ \\
\texttt{Choices:} $\mathcal{C}$ \\
\texttt{Answer:}
\end{prompt}

Using this prompt, the LLM must generate $a$, which is the highlighted text in the prompt box.

\subsection{Prompting Details} \label{appendix:prompt}

We design few-shot prompts following the format described by our prompt boxes. The few-shot examples were randomly selected from the training set, and we ensured that these contained a balanced distribution of output labels and that the demonstrations were shuffled. We created a few-shot prompt for each dataset. Both the UnifiedQA evaluation set and the contrast set used the exact same prompt. Even though this results in demonstrations with more than two choices, we found that this did not confuse models on the contrast set, as they never outputted an invalid letter (i.e. ``(C)'' when there are two choices). In the case of an invalid output, which stemmed from a non-letter choice, we marked the output as incorrect.

\section{Results}

\subsection{Qualitative Analysis Details} \label{appendix:qual_details}

\begin{figure*}
    \centering
    \fbox{\includegraphics[width=0.7\linewidth]{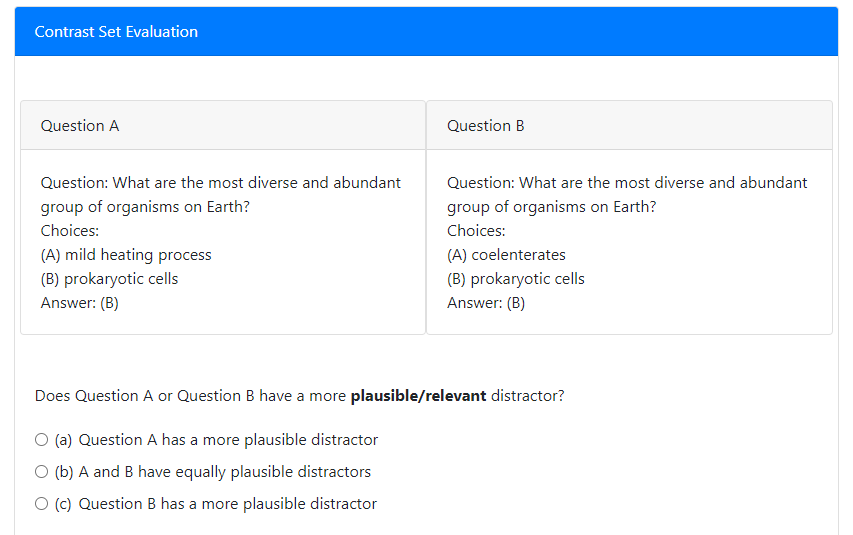}}
    \caption{Interface used by annotators.}
    \label{appendix:UI}
\end{figure*}

Below, we provide the exact instructions (Figure~\ref{appendix:instr}) and annotation interface (Figure~\ref{appendix:UI}) shown to our annotators. Our annotation interface is based on PrairieLearn \cite{west2015prairielearn}. Our use of plausibility and relevance for this annotation task is based on existing work \cite{gierl2017developing}.

The random baseline we compare against is the trivial solution described in \cref{subsection:graph_repr}. This baseline selects a random gold answer from the same dataset to form a set of choices. We apply this algorithm to the same 50 sampled instances as the ones annotators evaluated with our contrast set, meaning that the questions produced by this baseline only differ by the chosen distractor; the question, choices, and gold answer are all consistent across approaches.

\subsection{3-shot Prompting Results} \label{appendix_3_shot}

\begin{figure*}
    \centering
    \includegraphics[width=\linewidth]{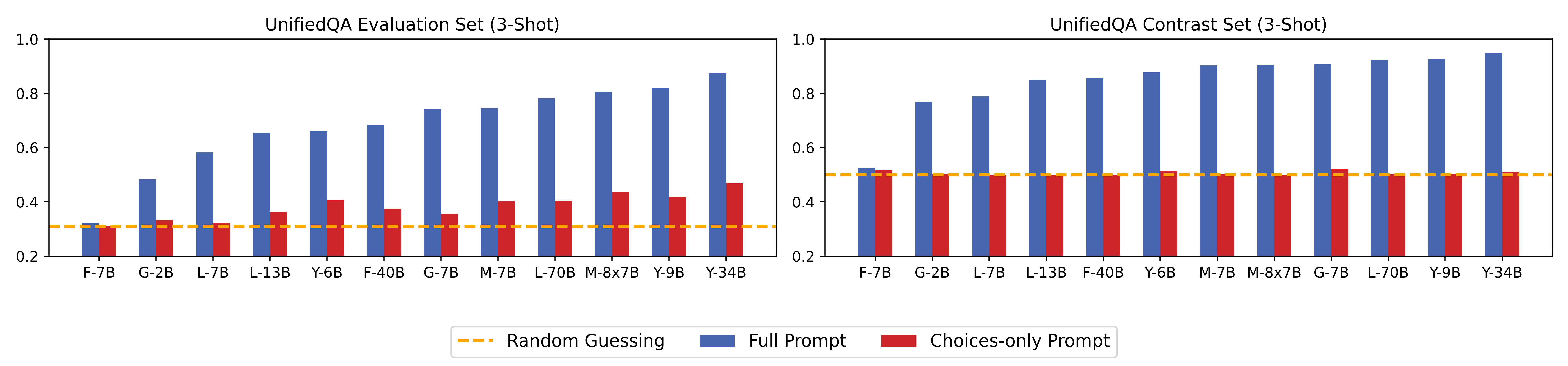}
    \caption{3-shot benchmarking of 12 LLMs on the UnifiedQA evaluation set and the contrast set, sorted by full-prompt accuracy. The same trends found for 5-shot and 10-shot prompting hold for 3-shot prompting.}
    \label{appendix:3_shot_graph}
\end{figure*}

In Figure~\ref{appendix:3_shot_graph}, we show the same results as Figure~\ref{fig:quant} but with three-shot prompting. The same trends of high choices-only accuracy and the consistency of full prompt rankings across evaluation and contrast sets both hold, with a Kendall's $\tau$ of 0.88. We did not test 0-shot prompting as we were working with base LLMs (i.e. unaligned and not instruction-tuned), which should not have the capability to complete tasks in a 0-shot manner. We believe that studying choices-only accuracy in 0-shot settings could be an interesting avenue for future work.

\subsection{UnifiedQA Evaluation Subset} \label{eval_subset}

\begin{figure*}
    \centering
    \includegraphics[width=\linewidth]{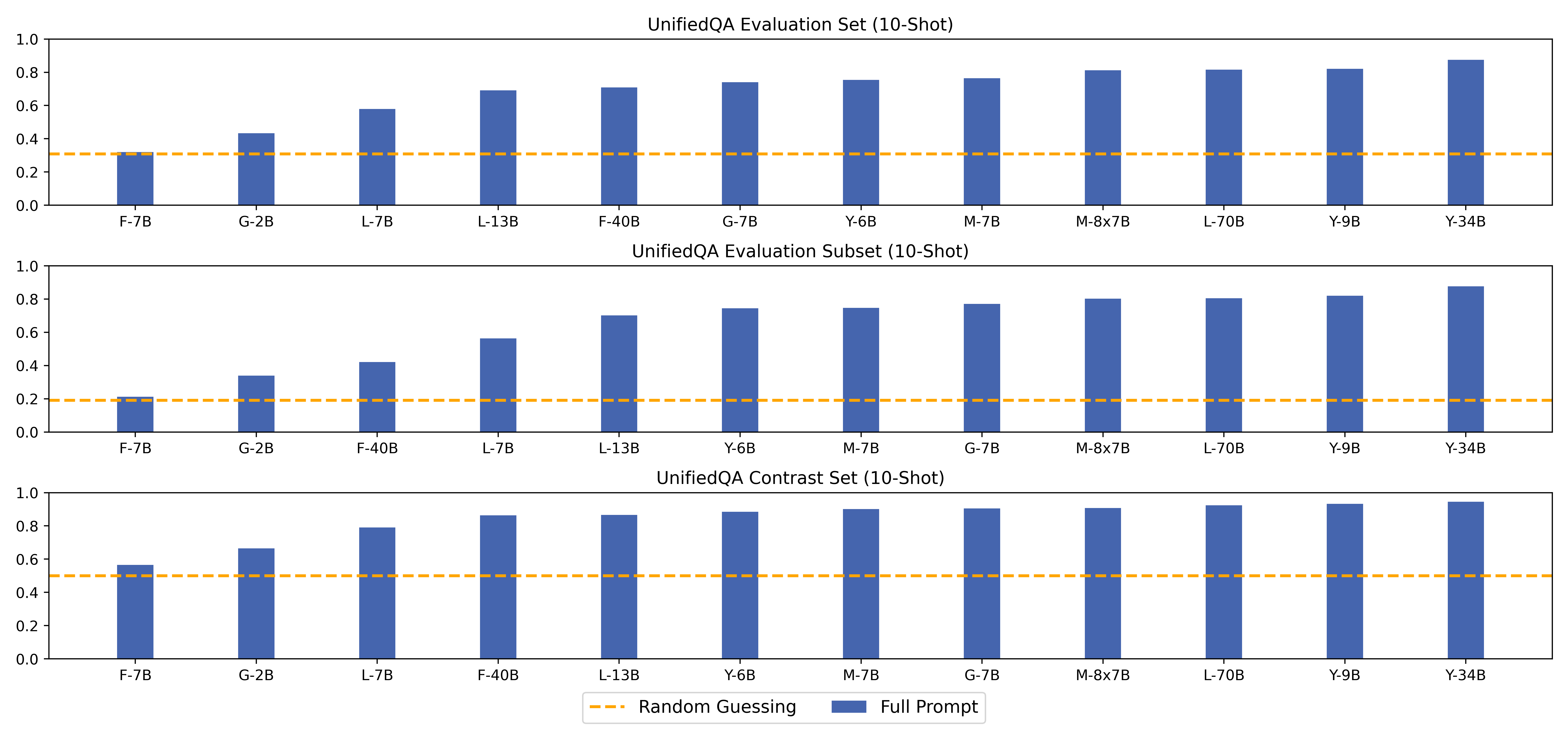}
    \caption{10-shot benchmarking of 12 LLMs on the UnifiedQA evaluation set, the contrast set, and the subset of the full UnifiedQA evaluation split with the same questions as the contrast set.}
    \label{appendix:extra_split}
\end{figure*}

Our mined contrast set only has two choices for every question, while the original evaluation set has questions ranging from 2 to 8 choices. To ensure the consistency of rankings is not confounded by the the reduction of possible choices, we also report the 10-shot accuracy on the subset of UnifiedQA that was used to derive the contrast set. This subset is essentially equivalent to the contrast set, but with additional choices on each question so that the number of choices are consistent. In Figure~\ref{appendix:extra_split}, the UnifiedQA Evaluation set and the UnifiedQA Evaluation subset have a similarly high consistency between rankings of full prompt accuracy. Thus, in our experiments, the number of choices on each question does not seem to largely influence the ranking of LLMs.

\begin{figure*}
    \centering
    \fbox{\includegraphics[width=0.7\linewidth]{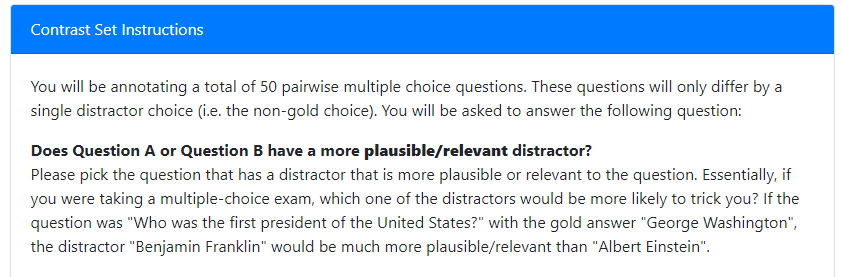}}
    \caption{Instructions shown to annotators.}
    \label{appendix:instr}
\end{figure*}

\end{document}